\documentclass{article} 
\usepackage{iclr2027_conference,times}

\usepackage{amsmath,amsfonts,bm}

\def\eqref#1{equation~\ref{#1}}

\def\1{\bm{1}}

\DeclareMathAlphabet{\mathsfit}{\encodingdefault}{\sfdefault}{m}{sl}
\SetMathAlphabet{\mathsfit}{bold}{\encodingdefault}{\sfdefault}{bx}{n}

\usepackage{graphicx}
\usepackage{hyperref}
\usepackage{url}
\usepackage{enumitem}
\usepackage{multirow}
\usepackage{booktabs}
\usepackage{colortbl}
\usepackage{subcaption}
\usepackage{pifont}
\usepackage{tcolorbox}
\usepackage{float}

\title{Vox-Infinity: Benchmarking the Limits of Long-Context Spoken Language Models}

\iclrfinalcopy
\author{Xize Cheng \thanks{equal contribution},\quad
Wenxu Jia$^*$,\quad
Chenyuhao Wen$^*$,\quad
Dongjie Fu$^*$\\
\textbf{Zehan Wang},\quad 
\textbf{Xinyu Zhang},\quad
\textbf{Tao Jin} \thanks{corresponding author}\\
Zhejiang University
}

\begin{document}

\maketitle

\begin{abstract}
Long-context understanding remains a fundamental challenge for large language models, as excessively long inputs often lead models to forget salient information. This issue is even more pronounced in the speech domain, where audio, as a low-compression modality, requires substantially more embeddings than text to preserve both semantic content and acoustic cues. To address this challenge, we introduce \textbf{Vox-Infinity}, the first benchmark specifically designed to evaluate long-context understanding in spoken language models. Vox-Infinity systematically extends audio history along two dimensions: turn count and turn duration. It covers a diverse range of representative scenarios with varying interaction structures and semantic complexity. Crucially, Vox-Infinity provides explicit answer-provenance annotations and organizes samples according to the amount of historical context required to resolve each query, enabling precise and length-aware evaluation. Extensive evaluations of seven representative spoken language models reveal a clear overall recency effect: models generally achieve higher accuracy when answer-supporting evidence is closer to the query, but struggle to retrieve and use evidence located farther back in the dialogue history. Cases and datasets are available at \url{https://vox-infinity.github.io}.
\end{abstract}

\section{Introduction}
Spoken language systems~\citep{kaplan2019siri,hachman2019microsoft,chu2023qwen,chu2024qwen2,ghosh2025audio,chen2024slam} aim to engage in intelligent and natural interactions with humans, requiring the ability to comprehend not only semantic content but also paralinguistic cues embedded in speech. Early systems~\citep{kaplan2019siri,hachman2019microsoft} primarily relied on automatic speech recognition (ASR)~\citep{yu2016automatic} to convert audio into text, and achieved basic dialogue functionality through intent recognition~\citep{liu2019review,zheng2017intent} and dialogue state tracking~\citep{williams2016dialog,jacqmin2022you} based on the transcribed content. With the emergence of large language models (LLM)~\citep{bai2023qwen,dubey2024llama,fang2025llama,yang2025qwen3}, some systems~\citep{speechteam2024funaudiollm,chen2025fireredchat,internlmxcomposer2_5_OL} have significantly improved in semantic understanding. Some recent approaches have integrated ASR and LLM to construct powerful multi-stage dialogue systems with strong interactive capabilities. However, as cascaded architectures, such systems~\citep{internlmxcomposer2_5_OL,chen2025fireredchat} still struggle to fully capture and interpret the rich information present in speech, particularly acoustic features such as intonation, emotion, and emphasis. To overcome these limitations and advance audio understanding, recent studies~\citep{tang2023salmonn,xu2025qwen2,chen2024slam,ghosh2025audio,zhang2024internlm} have proposed large audio language models that incorporate raw audio directly into the dialogue framework. This end-to-end integration allows models to better recognize and utilize nuanced acoustic signals~\citep{ao2024sd,cheng2025voxdialogue} that are often lost in ASR-based pipelines.

However, audio presents a significant challenge due to its low information density~\citep{ji2024wavchat}. Spoken content typically requires 12 to 25 embeddings~\citep{zeng2024glm,ji2024wavtokenizer} per second to encode, resulting in much longer input sequences than text for conveying equivalent semantic information. This substantially increases the effective context length in spoken language tasks and exacerbates the difficulty of long-context modeling. Prior research~\citep{bai2023longbench,ding2025kimi} on long-context language models has shown that performance tends to degrade as relevant information moves farther back in the input, a limitation that similarly affects spoken dialogue systems. While long-context understanding has been widely explored in textual~\citep{ding2024longrope} and visual modalities~\citep{chen2024longvila,zhang2024internlm}, research in the spoken language domain remains in its early stages. Most existing efforts~\citep{goel2025audio} focus on extending audio input beyond 30 seconds, a constraint largely imposed by the capacity of audio encoders such as Whisper~\citep{rouditchenko2024whisper}. However, a systematic evaluation of spoken language models under varying long-context conditions remains notably lacking.

To bridge this gap, we introduce \textsc{Vox-Infinity}, the first benchmark specifically designed to evaluate long-context understanding in spoken language models across varying context lengths. Vox-Infinity systematically extends audio history along two dimensions: the number of dialogue turns and the duration of each turn. As illustrated in Figure~\ref{fig:dialogue}, the benchmark encompasses three complementary scenarios: ultra multi-turn dialogues, personal monologues, and beyond-semantic dialogues.
To enable precise evaluation, each question is annotated with the exact locations of its answer-supporting evidence in the audio history. We define the required context duration as the temporal span from the earliest supporting evidence to the query. Unlike total session duration, this quantity captures how far the evidence required to answer a query lies in the dialogue history. Vox-Infinity groups test samples according to their required context duration, enabling fine-grained analysis of model performance across different evidence distances.
Using Vox-Infinity, we systematically evaluate seven representative spoken language models across diverse dialogue scenarios and required context durations. The results reveal a clear overall recency effect: models generally achieve higher accuracy when answer-supporting evidence is closer to the query, but struggle to retrieve and use evidence located farther back in the dialogue history. Vox-Infinity is available at \url{https://vox-infinity.github.io}. Our main contributions are as follows:

\begin{figure}
    \centering
    \begin{subfigure}[b]{0.32\linewidth}
        \includegraphics[width=\linewidth,trim=0 0 693 0,clip]{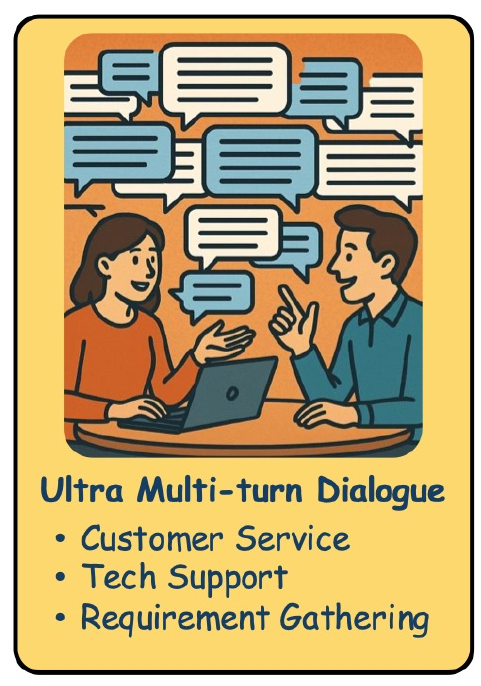}
    \end{subfigure}\hfill
    \begin{subfigure}[b]{0.32\linewidth}
        \includegraphics[width=\linewidth,trim=462 0 231 0,clip]{image/dialogue.pdf}
    \end{subfigure}\hfill
    \begin{subfigure}[b]{0.32\linewidth}
        \includegraphics[width=\linewidth,trim=693 0 0 0,clip]{image/dialogue.pdf}
    \end{subfigure}
    \caption{Illustration of three representative dialogue scenarios in \textsc{Vox-Infinity}. \textbf{(a) Ultra Multi-Turn Dialogue}: frequent back-and-forth exchanges requiring fine-grained information retrieval. \textbf{(b) Personal Monologue}: single-speaker long-form speech requiring reasoning over distributed evidence. \textbf{(c) Beyond-Semantic Dialogue}: extended conversations that require models to retain and retrieve emotions conveyed through vocal expression.}
    \label{fig:dialogue}
    \vspace{-0.7cm}
\end{figure}

\begin{itemize}[left=0em, itemsep=-0.5pt, label=\textbullet]
    \item We introduce Vox-Infinity, the first benchmark specifically designed to evaluate long-context understanding in spoken language models under varying context conditions. It systematically scales dialogue history along two axes, turn count and turn duration, and covers a diverse range of realistic long-context spoken language scenarios.
    
    \item Vox-Infinity provides explicit answer-provenance annotations and defines the required context duration of each query as the temporal span from the earliest supporting evidence to the query. This enables precise, length-sensitive evaluation across different evidence distances.
    
    \item We systematically evaluate seven representative spoken language models and reveal a clear overall recency effect. Models generally perform better when answer-supporting evidence is closer to the query and struggle to retrieve and use evidence located farther back in the dialogue history.
\end{itemize}

\section{Related Works}

\subsection{Spoken Dialogue Systems}

The rapid advancement of large language models (LLMs) has fueled the emergence of increasingly powerful spoken dialogue systems. A notable early milestone was SpeechGPT~\citep{zhang2023speechgpt}, the first speech-centric LLM to integrate discrete speech units into a unified language modeling framework. Building on this foundation, large-scale audio language models such as Qwen-Audio 1/2~\citep{chu2023qwen,chu2024qwen2} extended capabilities to over 30 audio-related tasks, including speech recognition, speech translation, and audio event detection—laying the groundwork for more sophisticated spoken language understanding.
These core audio modeling capabilities have since enabled the development of specialized dialogue systems. For example, StyleTalk~\citep{lin2024advancing} focuses on emotional dialogue, introducing the first model capable of generating speech with distinct emotional prosody. Models such as GLM-4-Voice~\citep{zeng2024glm}, LLaMA-Omni 1/2~\citep{fang2024llama,fang2025llama}, and the Qwen-Omni series~\citep{xu2025qwen2,xu2025qwen3omni,qwen2026qwen35omni} support end-to-end spoken interaction through unified speech understanding and generation.

Despite these advancements, most current systems remain constrained by limited context windows. Because audio is a low-compression modality, representing equivalent semantic content requires much longer sequences than text, posing serious challenges for long-context modeling. To address this challenge, the Audio Flamingo series~\citep{ghosh2025audio,goel2025audio,ghosh2026audioflamingonext} has progressively extended support for long-form audio understanding, with its latest version supporting complex audio inputs of up to 30 minutes. In parallel, models such as Slam-Omni~\citep{chen2024slam} and Step-Audio2~\citep{wu2025step} alleviated audio compression bottlenecks by incorporating text-based history representations, which significantly improve memory efficiency and dialogue history retention. MiMo-Audio~\citep{coreteam2025mimoaudio} adopts a complementary approach by aggregating every four consecutive 25-Hz tokenizer frames into a 6.25-Hz representation for the LLM, thereby reducing the length of audio sequences.

\subsection{Long-Context Modeling and Benchmarks}
Long-context modeling~\citep{li2024long} has become a central research area in the development of LLMs, as extended inputs are essential for maintaining discourse coherence and retrieving information introduced earlier in the context. Prior work has expanded model context windows through architectural modifications, positional extrapolation, and context compression~\citep{ding2024longrope,jiang2023longllmlingua}. More recent work has further scaled context modeling through sparse and linear attention mechanisms. Qwen2.5-1M~\citep{yang2025qwen251m} combines long-context training, length extrapolation, and sparse attention to support contexts of up to one million tokens. Kimi Linear~\citep{kimiteam2025kimilinear} introduces Kimi Delta Attention for efficient inference at the million-token scale, while Kimi K3~\citep{kimiteam2026kimik3} further scales this architecture with a one-million-token context window. However, a larger nominal context window does not necessarily imply that a model can reliably use information throughout the input. Prior studies have shown that model performance can vary substantially with the position of relevant evidence~\citep{liu2024lost}.

To systematically evaluate whether models can effectively use extended contexts, a number of long-context benchmarks have been developed. Early efforts such as LAMBADA~\citep{paperno2016lambada} and Long Range Arena~\citep{tay2020long} focused on controlled settings for testing long-range dependencies. More comprehensive benchmarks, including LongBench~\citep{bai2023longbench} and L-Eval~\citep{an2023eval}, expanded evaluation to summarization, question answering, reasoning, and dialogue. Subsequent benchmarks further increased context scale and task difficulty: InfiniteBench~\citep{zhang2024infty} extends evaluation beyond 100K tokens, LongBench-V2~\citep{bai2024longbench} emphasizes realistic and challenging long-context tasks, and RULER~\citep{hsieh2024ruler} evaluates models under configurable sequence lengths and task complexities. Together, these studies show that nominal context size alone is insufficient to characterize long-context capability, as performance can vary with sequence length, task complexity, and evidence position.

Despite these advancements, existing evaluations remain almost exclusively text-only, overlooking the distinctive challenges of spoken language, such as acoustic cues~\citep{ao2024sd}, and significantly longer sequence lengths due to audio’s low information density~\citep{ji2024wavchat}. These factors make spoken language understanding a fundamentally different long-context problem. To address this gap, we propose Vox-Infinity, the first benchmark explicitly designed to evaluate long-context reasoning in spoken language models. Vox-Infinity scales audio history along two axes (turn count and duration) and incorporates a diverse set of realistic spoken scenarios. By providing detailed annotations and organizing test cases by answer provenance and context length, Vox-Infinity enables fine-grained context-aware evaluation, pushing the boundaries of long-context modeling beyond the text domain to better reflect real-world usage of spoken language systems.

\begin{figure}[tb]
    \centering
    \begin{subfigure}[b]{0.32\linewidth}
        \centering
        \includegraphics[width=\linewidth]{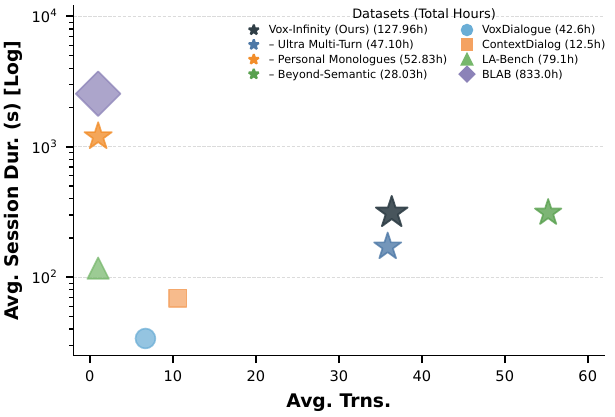}
        \caption{Dataset scale comparison.}
        \label{fig:scale}
    \end{subfigure}
    \hfill
    \begin{subfigure}[b]{0.34\linewidth}
        \centering
        \includegraphics[width=\linewidth]{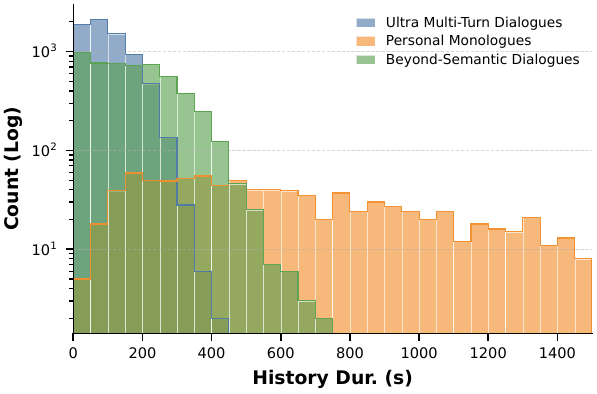}
        \caption{Required-history-duration distribution.}
        \label{fig:history_duration_histogram}
    \end{subfigure}
    \hfill
    \begin{subfigure}[b]{0.32\linewidth}
        \centering
        \includegraphics[width=\linewidth]{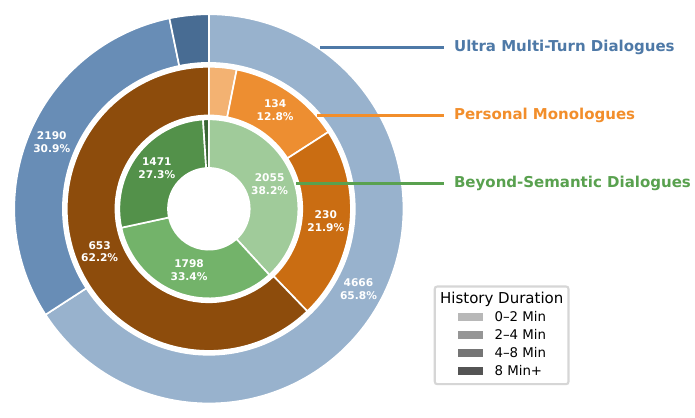}
        \caption{Proportion of QA samples by required history duration.}
        \label{fig:duration}
    \end{subfigure}
    \hfill
    \begin{subfigure}[b]{\linewidth}
        \centering
        \includegraphics[width=\linewidth]{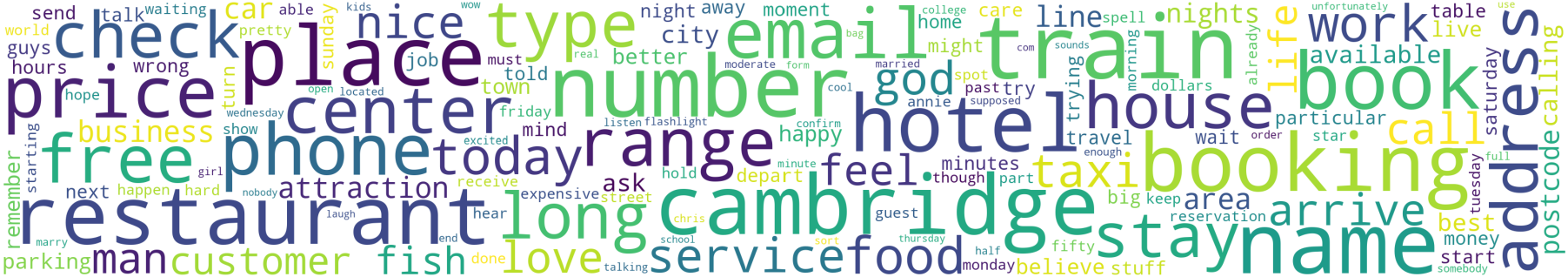}
        \caption{Word cloud visualization of common topics.}
        \label{fig:wordcloud}
    \end{subfigure}
    \vspace{-0.5cm}
    \caption{Statistics and characteristics of \textsc{Vox-Infinity}. \textbf{(a)} Dataset-scale comparison by average turns per session and average complete session duration; marker size encodes total hours, and stars denote \textsc{Vox-Infinity} and its three subsets. \textbf{(b)} Distribution of required history durations using 50-second bins and a logarithmic count axis. \textbf{(c)} Percentage of QA samples in the 0--2, 2--4, 4--8, and 8+ minute required-history-duration bins. \textbf{(d)} Word cloud of common topics.}
    \label{fig:stastic}
    \vspace{-0.5cm}
\end{figure}


\section{Vox-Infinity: Long-context Spoken Language Benchmark}

\subsection{Dataset Construction}

\textbf{Stage 1: Long-Context Spoken Dialogue Generation.}  
\textsc{Vox-Infinity} encompasses three complementary spoken dialogue scenarios. 
\textit{\textbf{(I) Ultra Multi-Turn Dialogues}} consist of conversations with more than 20 turns, constructed from SpokenWOZ~\citep{si2024spokenwoz}, a task-oriented dialogue dataset recorded by real speakers. Questions in this subset evaluate whether models can retrieve relevant information from dense, long-turn dialogue histories.
\textit{\textbf{(II) Personal Monologues}} feature single-speaker speeches longer than two minutes, derived from GigaSpeech~\citep{chen2021gigaspeech}. Based on timestamp annotations, we reconstruct short segments into their original long-form monologues. The corresponding questions require multi-hop reasoning over evidence distributed throughout the long-form speech.
\textit{\textbf{(III) Beyond-Semantic Dialogues}} evaluate whether models can retain paralinguistic emotion cues over extended dialogue histories. We construct this subset from IEMOCAP~\citep{busso2008iemocap}. After isolating the target speaker using the MossFormer2 speech separation model~\citep{zhao2024mossformer2}, we associate each turn with its fine-grained emotion annotation. However, emotions in naturally occurring speech can often be inferred from lexical and semantic content alone, allowing models to answer correctly without relying on acoustic cues. To disentangle acoustic emotion recognition from text-based semantic inference, we additionally select neutral turns and use Qwen-TTS to synthesize the same textual content in different emotional styles. The synthesized turns are independently validated by Gemini and Qwen-Omni, and only samples for which both models recognize the intended emotion are retained. By combining naturally emotional speech with semantically controlled synthetic speech, this subset evaluates whether models can retrieve and recognize emotional cues expressed in distant dialogue turns.

\textbf{Stage 2: Question–Answer Pair Generation.}
In the second stage, we generate multiple question-answer (QA) pairs for each session to evaluate fine-grained retrieval, multi-hop reasoning, and paralinguistic memory across varying required context durations. Since model responses may evolve as the dialogue progresses, we employ GPT-4 to generate diverse questions based on the whole dialogue to ensure each yields a unique and unambiguous answer. To precisely control the required context, GPT-4 is also instructed to specify which dialogue turns must be referenced when answering each question. Detailed prompt design and generation strategies are provided in the appendix~\ref{App:QA}.

\textbf{Stage 3: Data Verification and Quality Control.}
Finally, to ensure benchmark quality, each QA pair is manually examined to verify that its answer is correct and unique within the corresponding dialogue context. The associated answer-provenance annotation is further checked to confirm that the annotated turns provide the supporting evidence required to answer the question. Samples with ambiguous answers, factual inconsistencies, or inaccurate provenance annotations are regenerated and subjected to another round of manual verification. This process yields a reliable long-context spoken language benchmark covering diverse scenarios and capabilities.

\begin{table}[tb]
\small
\setlength{\tabcolsep}{2pt}
\caption{ Comparison of existing long-context benchmarks across textual and audio modalities. \textbf{Avg. Tok.} denotes the average number of tokens per session, 
\textbf{Avg.Dur.} denotes the average duration per session, 
and \textbf{Avg.Trns.} denotes the average number of turns per session. 
\textbf{Prov.} indicates whether answer provenance annotations are available. 
\textbf{Length distribution} refers to the context-length ranges used for evaluation; 
$^\ddagger$ for datasets without explicit definitions, we adopt the near-upper quartile of context length as the upper bound. 
For audio benchmarks, average duration is calculated as total duration divided by the number of sessions; 
$^\dagger$ token counts are estimated assuming 25 tokens per second.
}
\label{tab:compare}
\centering
\begin{tabular}{lrrrrrrcl}
\toprule
\multirow{2}{*}{\textbf{\small Benchmarks}} 
& \multirow{2}{*}{\textbf{\small Session}} 
& \multirow{2}{*}{\textbf{\small QAs}} 
& \multirow{2}{*}{\textbf{\small Dur.(h)}}
& \multirow{2}{*}{\textbf{\begin{tabular}[c]{@{}c@{}}Avg.\\ Tok.$^\dagger$\end{tabular}}} 
& \multirow{2}{*}{\textbf{\begin{tabular}[c]{@{}c@{}}Avg.\\ Dur.(s)\end{tabular}}} 
& \multirow{2}{*}{\textbf{\begin{tabular}[c]{@{}c@{}}Avg.\\ Trns.\end{tabular}}} 
& \multirow{2}{*}{\textbf{\small Prov.}} 
& \multirow{2}{*}{\textbf{Length Distribution$^\ddagger$}} \\
                                     &                                    &                            &                            &                            &                                                                                 &                                               \\
\midrule
\multicolumn{9}{l}{\cellcolor[HTML]{EFEFEF}\textit{Long Context Benchmark (Textual)}}  \\
\scriptsize{LongBench}\scriptsize{~\citep{bai2023longbench}}      
& 4,550
& 4,550
& -
& 8.1k
& -
& 1.00
& \textcolor[HTML]{FF0000}{\ding{55}}
& \scriptsize{0-4k, 4k-8k, 8k+}                              \\
\scriptsize{$\infty$ BENCH}\scriptsize{~\citep{zhang2024infty}}
& 3,946
& 3,946
& -
& 200k
& -
& 1.00
& \textcolor[HTML]{FF0000}{\ding{55}}
& \scriptsize{0-200k}                               \\
\scriptsize{LongBench-V2}\scriptsize{~\citep{bai2024longbench}}               
& 503
& 503      
& -
& 54k
& -                   
& 1.00  
& \textcolor[HTML]{FF0000}{\ding{55}}                  
& \scriptsize{0-32k, 32k-128k, 128k+}                           \\ \midrule
\multicolumn{9}{l}{\cellcolor[HTML]{EFEFEF}\textit{Long Context Benchmark (Audio)}}  \\
\scriptsize{VoxDialogue~\citep{cheng2025voxdialogue}}     
& 4,526                            
& 4,526
& 42.56
& 0.8k
& 33.9
& 6.70
& \textcolor[HTML]{FF0000}{\ding{55}}
& \scriptsize{0-2min}                              \\ 
\scriptsize{ContextDialog~\citep{kim2025does}}     
& 653                            
& 2,612
& 12.5
& 1.7k
& 68.9
& 10.60
& \textcolor[HTML]{FF0000}{\ding{55}}
& \scriptsize{0-2min}                              \\ 
\scriptsize{LA-Bench~\citep{kong2024audio}}     
& 2,429
& 2,429
& 79.12
& 2.9k
& 117.3
& 1.00
& \textcolor[HTML]{FF0000}{\ding{55}}
& \scriptsize{0-3min+}                              \\ 
\scriptsize{BLAB~\citep{ahia2025blab}}
& 1,176
& 1,600
& 833
& 63.8k
& 2,550.0
& 1.00
& \textcolor[HTML]{FF0000}{\ding{55}}
& \scriptsize{0-60min+}                              \\ 
\midrule
\scriptsize{Vox-Infinity (ours)}        
& 1,474
& 13,516
& 127.96
& 7.8k
& 312.5
& 36.38
& \textcolor[HTML]{00B050}{\ding{51}}
& \scriptsize{0-2min, 2-4min, 4-8min, 8min+}   
\\
\textit{\quad- \scriptsize{Ultra Multi-Turn Dialogue}}     
& 991
& 7,086
& 47.10
& 4.3k
& 171.1
& 35.89
& \textcolor[HTML]{00B050}{\ding{51}}
& \scriptsize{0-2min, 2-4min, 4-8min}
\\
\textit{\quad- \scriptsize{Personal Monologues}}          
& 159
& 1,050
& 52.83
& 29.9k
& 1,196.1
& 1.00
& \textcolor[HTML]{00B050}{\ding{51}}
& \scriptsize{0-2min, 2-4min, 4-8min, 8min+}\\
\textit{\quad- \scriptsize{Beyond-Semantic Dialogue}}       
& 324
& 5,380
& 28.03
& 7.8k
& 311.4
& 55.23
& \textcolor[HTML]{00B050}{\ding{51}}
& \scriptsize{0-2min, 2-4min, 4-8min, 8min+}
\\
\bottomrule
\end{tabular}
\vspace{-0.5cm}
\end{table}

\subsection{Dataset Statistics}

\textbf{Detailed Statistics of Vox-Infinity.}
To provide a comprehensive overview of Vox-Infinity, Figure~\ref{fig:stastic} presents detailed dataset statistics.
As shown in Figure~\ref{fig:scale}, we compare the three subsets of Vox-Infinity with existing long-audio benchmarks in terms of average turns per session and average complete session duration. Across its three subsets, Vox-Infinity contains 1,474 sessions, 13,516 QA pairs, and 127.96 hours of audio, with an average session duration of 312.5 seconds. Beyond-Semantic contains 324 sessions and 5,380 QA pairs, with an average of 55.23 turns and 311.4 seconds per session. Together, the three subsets cover complementary regimes ranging from dense multi-turn interactions to long single-speaker monologues.
Figure~\ref{fig:history_duration_histogram} illustrates the distribution of required history durations across the three subsets. Figure~\ref{fig:duration} summarizes the same QA-level distributions using the evaluation bins adopted throughout the paper. Ultra Multi-Turn is concentrated below four minutes, Personal Monologues contain predominantly eight-minute-plus dependencies, and Beyond-Semantic spans all four required-history-duration ranges.
Finally, Figure~\ref{fig:wordcloud} presents a word cloud of common topics in Vox-Infinity. The dataset features a high concentration of queries targeting key contextual information—such as phone numbers, train IDs, and restaurants—all of which have explicit answers, making them ideal for assessing a model’s ability to retain and utilize long-term contextual memory. Beyond-Semantic additionally introduces long-range paralinguistic emotion cues, emphasizing challenges that cannot be captured by text-only dialogue histories.

\textbf{Comparison with Existing Benchmarks.} 
Table~\ref{tab:compare} situates Vox-Infinity within the landscape of long-context benchmarks across text and audio modalities. Text-based benchmarks such as LongBench~\citep{bai2023longbench, bai2024longbench} enable performance evaluation under varying context lengths. In contrast, existing audio benchmarks~\citep{goel2025audio} primarily aim to overcome the 30-second input limit of encoders like Whisper~\citep{radford2023robust}, focusing on extending input length rather than analyzing performance across different context regimes.
Vox-Infinity fills this gap by being the first audio benchmark to explicitly evaluate models under multiple context-length conditions, allowing fine-grained analysis of long-context reasoning. Compared with benchmarks like BLAB~\citep{ahia2025blab}, whose recordings span 15 minutes to 2 hours, Vox-Infinity adopts more practical dialogue durations while preserving contextual diversity. This makes it better suited for realistic and scalable spoken dialogue evaluation.

Another limitation of prior QA-based benchmarks~\citep{bai2024longbench,goel2025audio} is the absence of answer provenance annotations, which makes the true amount of context required to answer a question ambiguous. Vox-Infinity addresses this issue by explicitly labeling provenance, enabling precise measurement of the actual context length needed to support each answer. By jointly considering context-length granularity and answer provenance, Vox-Infinity establishes a benchmark that is more analytically rigorous for long-context modeling.

\subsection{Evaluation Suite}
Following the evaluation suite of LongAudioBench~\citep{ghosh2025audio}, we adopt Accuracy as the primary evaluation metric, which measures the proportion of correctly answered samples among all test cases. Given the high diversity of responses generated by Spoken Dialogue Systems, we follow prior work~\citep{duan2024vlmevalkit,ghosh2025audio} and employ GPT-4o~\citep{gpt4o} to assess the model outputs. Responses are categorized into three labels: Correct, Incorrect, and Unknown. The Unknown label allows GPT-4o to indicate that a response cannot be reliably judged as correct or incorrect, avoiding forced binary decisions in ambiguous cases. Detailed prompt templates for this evaluation process are provided in Appendix~\ref{app:prompt_template}.

\section{Benchmarking Long-Context Spoken Language Models}

\begin{table}[tb]
\caption{Accuracy of spoken language models across dialogue scenarios and required-context-duration bins. \textit{Beyond-Semantic} uses multiple-choice questions, whereas \textit{Ultra Multi-Turn} and \textit{Personal Monologues} use open-ended questions. Here, \textit{m} denotes minutes and \textit{Avg.} denotes the average accuracy within each subset. All values are percentages, and $^\dagger$ marks closed-source models.}
\label{tab:main}
\vspace{-0.3cm}
\centering
\resizebox{\textwidth}{!}{%
\begin{tabular}{@{}l*{14}{c}@{}}
\toprule
\multicolumn{1}{c}{}
& \multicolumn{4}{c}{\textit{\textbf{Ultra Multi-Turn}}}
& \multicolumn{5}{c}{\textit{\textbf{Personal Monologues}}}
& \multicolumn{5}{c}{\textit{\textbf{Beyond-Semantic}}} \\
\cmidrule(lr){2-5}\cmidrule(lr){6-10}\cmidrule(lr){11-15}
\multicolumn{1}{c}{\multirow{-2}{*}{\textbf{Model}}}
& \textit{\textbf{0--2m}}
& \textit{\textbf{2--4m}}
& \textit{\textbf{4--8m}}
& \textit{\textbf{Avg.}}
& \textit{\textbf{0--2m}}
& \textit{\textbf{2--4m}}
& \textit{\textbf{4--8m}}
& \textit{\textbf{8m+}}
& \textit{\textbf{Avg.}}
& \textit{\textbf{0--2m}}
& \textit{\textbf{2--4m}}
& \textit{\textbf{4--8m}}
& \textit{\textbf{8m+}}
& \textit{\textbf{Avg.}} \\
\midrule
\multicolumn{15}{l}{\cellcolor[HTML]{EFEFEF}\textit{Text-as-history}} \\
Qwen3-Omni
& 85.5 & 84.2 & 77.0 & 84.8
& 84.2 & 86.8 & 84.7 & 85.5 & 85.5
& 49.5 & 46.2 & 42.4 & 25.0 & 46.2 \\
GLM-4-Voice
& 61.4 & 56.3 & 48.3 & 59.4
& 27.3 & 36.6 & 26.1 & 26.3 & 27.6
& 43.5 & 42.4 & 37.0 & 10.7 & 41.0 \\
AudioFlamingo3
& 45.6 & 40.7 & 23.0 & 43.4
& 42.4 & 58.2 & 47.8 & 47.8 & 49.0
& 20.0 & 20.1 & 15.0 & 8.9 & 18.5 \\
AudioFlamingo Next
& 76.1 & 73.2 & 67.0 & 74.9
& 45.5 & 62.7 & 63.0 & 62.6 & 62.2
& 46.7 & 40.6 & 35.1 & 16.1 & 41.2 \\
MiMo-Audio
& 76.1 & 78.1 & 70.9 & 76.6
& 63.6 & 79.9 & 74.3 & 63.1 & 67.7
& 32.5 & 30.2 & 26.8 & 16.1 & 30.0 \\
$^\dagger$Gemini-3.1-Pro
& 80.3 & 77.6 & 66.1 & 79.0
& 45.5 & 81.3 & 80.9 & 85.3 & 82.6
& 58.2 & 54.2 & 46.4 & 19.6 & 53.2 \\
$^\dagger$Qwen3.5-Plus
& 82.2 & 79.7 & 71.3 & 81.1
& 57.6 & 81.3 & 84.3 & 86.4 & 84.4
& 57.0 & 52.9 & 48.4 & 17.9 & 52.9 \\
\midrule
\multicolumn{15}{l}{\cellcolor[HTML]{EFEFEF}\textit{Audio-as-history}} \\
Qwen3-Omni
& 56.0 & 51.9 & 66.7 & 55.0
& 50.0 & 75.8 & 72.9 & 69.6 & 70.5
& 56.6 & 53.1 & 49.4 & 32.1 & 53.2 \\
GLM-4-Voice
& 36.9 & 31.5 & 23.5 & 34.8
& 12.1 & 20.1 & 19.1 & 9.8 & 13.2
& 24.1 & 23.2 & 19.2 & 8.9 & 22.3 \\
AudioFlamingo3
& 35.4 & 34.3 & 29.2 & 34.4
& 60.6 & 64.9 & 49.1 & 37.1 & 44.0
& 26.6 & 26.8 & 23.0 & 12.5 & 25.5 \\
AudioFlamingo Next
& 30.8 & 35.3 & 26.1 & 32.2
& 45.5 & 51.5 & 49.1 & 42.9 & 45.4
& 50.9 & 46.7 & 42.2 & 20.0 & 46.7 \\
MiMo-Audio
& 24.6 & 19.4 & 13.8 & 22.5
& 25.0 & 6.3 & 22.1 & 10.8 & 12.5
& 42.7 & 38.4 & 37.6 & 21.4 & 40.3 \\
$^\dagger$Gemini-3.1-Pro
& 63.2 & 62.6 & 54.8 & 62.8
& 33.3 & 62.7 & 60.4 & 60.8 & 60.1
& 62.7 & 57.7 & 51.0 & 31.9 & 57.5 \\
$^\dagger$Qwen3.5-Plus
& 57.2 & 56.6 & 53.0 & 56.8
& 39.4 & 64.9 & 64.8 & 65.7 & 64.6
& 60.8 & 57.8 & 55.3 & 39.3 & 58.1 \\
\bottomrule
\end{tabular}%
}
\vspace{-0.5cm}
\end{table}

\subsection{Benchmarking SLMs Across Context Lengths}

\subsubsection{Comparison Models.}

We evaluate seven representative spoken language models: Qwen3-Omni, GLM-4-Voice~\citep{zeng2024glm}, AudioFlamingo3~\citep{goel2025audio}, AudioFlamingo Next, MiMo-Audio~\citep{coreteam2025mimoaudio}, Gemini-3.1-Pro, and Qwen3.5-Plus. Each model is evaluated with dialogue history represented entirely as text or audio. Text histories are transcribed using the open-source SenseVoice model~\citep{speechteam2024funaudiollm}, and all models use a maximum context window of 32,768 tokens.

\subsubsection{Main Results}
\begin{figure*}[t]
    \centering
    \includegraphics[width=0.82\textwidth]{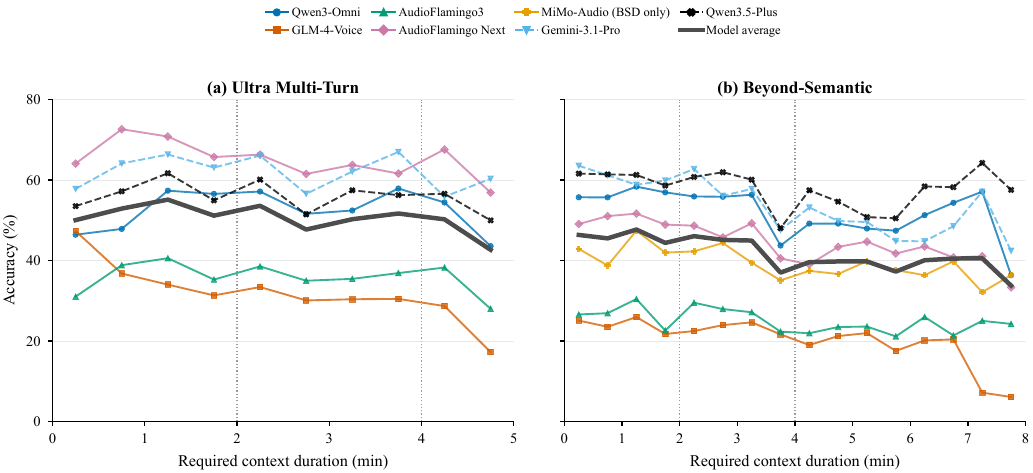}
    \caption{Fine-grained accuracy under audio history across 30-second required-context-duration intervals on \textbf{(a)} Ultra Multi-Turn and \textbf{(b)} Beyond-Semantic Dialogues. Thin curves represent individual models, while thick gray curves show their mean within each interval. Intervals containing fewer than 20 examples are omitted, and MiMo-Audio is included only in \textbf{(b)}. Despite local fluctuations, both subsets exhibit an overall recency effect.}
    \label{fig:recency_curves}
\end{figure*}

Table~\ref{tab:main} compares model performance across the three subsets of Vox-Infinity. As the required context duration increases, models using either text or audio history generally exhibit performance degradation. To provide a finer-grained view of this trend, Figure~\ref{fig:recency_curves} plots model accuracy under audio history across 30-second intervals of required context duration. Although the resulting curves are not strictly monotonic because each interval contains different questions, the model-averaged curves reveal an overall recency effect. Consistent with Table~\ref{tab:main}, the average accuracy across all seven models decreases from 43.4\% to 38.2\% on Ultra Multi-Turn Dialogues between the 0--2 min and 4--8 min ranges. For Beyond-Semantic Dialogues, the corresponding average falls from 46.3\% to 39.7\% and decreases further to 23.7\% beyond eight minutes. These results indicate that distant evidence challenges both fine-grained semantic information retrieval and paralinguistic memory, with the latter exhibiting greater degradation over extended histories.

Across tasks that primarily depend on linguistic content, models generally perform better with text history, with the clearest advantages observed on Ultra Multi-Turn Dialogues and Personal Monologues. This pattern suggests that compact transcripts support the retention and use of long-range semantic information more effectively than substantially longer audio sequences. The gap is especially pronounced on Ultra Multi-Turn Dialogues, where questions often require recovering exact slot values, such as phone numbers and train identifiers, from densely interleaved turns. Text history outperforms audio history for every evaluated model on this subset, with mean accuracies of 71.3\% and 42.6\%, respectively. This substantial difference indicates that current spoken language models struggle to retain and recover precise lexical evidence from long audio histories.

However, this advantage does not extend to queries that depend on paralinguistic evidence. Transcripts preserve linguistic content but discard prosodic and vocal cues, including intonation, rhythm, and intensity. Accordingly, audio history outperforms text history for six of the seven evaluated models on Beyond-Semantic Dialogues. For example, MiMo-Audio achieves 30.0\% with text history and 40.3\% with audio history. These results reveal complementary strengths between the two history representations. Text history more effectively supports the retention and retrieval of long-range semantic information, whereas audio history retains paralinguistic evidence that is unavailable in transcripts. Taken together, the three subsets expose distinct limitations in fine-grained retrieval, multi-hop reasoning, and paralinguistic memory, demonstrating the diagnostic value of Vox-Infinity beyond a single aggregate score.

\subsubsection{Disentangling Acoustic Emotion from Semantic Cues}
To distinguish acoustic emotion recognition from semantic inference, we divide Beyond-Semantic Dialogues into natural and semantically controlled synthetic samples. Natural samples retain the original IEMOCAP recordings, in which emotion may be conveyed by both lexical content and vocal expression. Controlled samples instead use semantically neutral utterances synthesized in designated emotional styles, making the intended emotion difficult to infer from text alone. Table~\ref{tab:emotion_control} compares text and audio histories under both settings.

\begin{table}[H]
\centering
\caption{Accuracy on natural ($n=2{,}758$) and semantically controlled ($n=2{,}622$) samples from Beyond-Semantic Dialogues. Text and Audio denote the history modality, and $\Delta$ reports Audio minus Text in percentage points. All values are percentages.}
\label{tab:emotion_control}
\begin{tabular}{lrrrrrr}
\toprule
& \multicolumn{3}{c}{\textit{Natural}}
& \multicolumn{3}{c}{\textit{Semantically Controlled}} \\
\cmidrule(lr){2-4}\cmidrule(lr){5-7}
\textbf{Model}
& \textbf{Text}
& \textbf{Audio}
& $\boldsymbol{\Delta}$
& \textbf{Text}
& \textbf{Audio}
& $\boldsymbol{\Delta}$ \\
\midrule
Qwen3-Omni                & 65.3 & 68.3 & +3.0 & 26.1 & 37.3 & +11.2 \\
GLM-4-Voice               & 52.9 & 28.4 & -24.6 & 28.5 & 16.0 & -12.5 \\
AudioFlamingo3            & 28.9 & 35.1 & +6.1 & 7.6 & 15.4 & +7.9 \\
AudioFlamingo Next        & 57.8 & 57.9 & +0.1 & 23.6 & 34.9 & +11.3 \\
MiMo-Audio                & 41.4 & 53.8 & +12.4 & 18.0 & 26.0 & +8.1 \\
$^\dagger$Gemini-3.1-Pro  & 74.0 & 74.9 & +0.9 & 31.4 & 39.2 & +7.9 \\
$^\dagger$Qwen3.5-Plus    & 73.6 & 71.4 & -2.2 & 31.2 & 44.1 & +12.9 \\
\bottomrule
\end{tabular}
\end{table}

For most models, audio history outperforms text history, and this advantage becomes substantially clearer when semantic cues are controlled. On natural samples, audio history performs better for five of the seven models, but the model-averaged accuracies of audio and text histories are nearly identical at 55.7\% and 56.3\%, respectively. This small gap is consistent with many natural utterances containing lexical evidence of emotion, which text history can preserve and retrieve effectively over long contexts. On controlled samples, audio history outperforms text history for six of the seven models and raises model-averaged accuracy from 23.8\% to 30.4\%, a gain of 6.6 percentage points. Because their lexical content is emotionally neutral, these samples require models to rely more heavily on prosody and vocal intensity, such as the raised volume and sharp intonation associated with anger, to retain and recover the intended emotion. Since the controlled samples contain synthesized speech, we interpret within-group modality gaps rather than cross-group differences in absolute accuracy.

GLM-4-Voice is the main exception, with text history outperforming audio history in both groups. A likely reason is that its discrete speech tokenizer and semantically oriented training pipeline prioritize linguistic content while retaining limited emotion-related prosody~\citep{zeng2024glm}. This exception highlights that the benefit of audio history depends on whether the underlying speech representation preserves paralinguistic information.

\subsubsection{Human Validation of LLM-Based Evaluation}
To assess the reliability of the LLM judge, we randomly sample 300 audio-history model outputs from Qwen3-Omni and Audio Flamingo Next, comprising 50 examples from each of the three subsets for each model. Beyond-Semantic samples are drawn from the combined natural and semantically controlled subsets. All outputs are sampled without replacement using a fixed random seed (20260910) and subsequently reviewed by human annotators. Table~\ref{tab:human} reports both the resulting model accuracies and the agreement between LLM-based and human judgments. Because this analysis uses a random sample rather than the complete benchmark, the sample-level accuracies may differ from those in Table~\ref{tab:main}.

\begin{table}[H]
\centering
\caption{Agreement between LLM-based evaluation and human judgments on 300 randomly sampled audio-history outputs. Each subset cell contains 50 examples. Agreement denotes the percentage of matching labels.}
\label{tab:human}
\setlength{\tabcolsep}{3pt}
\renewcommand{\arraystretch}{0.86}
\begin{tabular}{@{}l|cccc@{}}
\toprule
\textbf{Method} & \textbf{Ultra Multi-Turn} & \textbf{Personal Monologues} & \textbf{Beyond-Semantic} & \textbf{Overall} \\
\midrule
\multicolumn{5}{l}{\cellcolor[HTML]{EFEFEF}\textit{Qwen3-Omni}} \\
LLM-eval & 48.0 & 74.0 & 54.0 & 58.7 \\
Human-eval & 46.0 & 76.0 & 54.0 & 58.7 \\
Agreement & 98.0 & 98.0 & 100.0 & 98.7 \\
\multicolumn{5}{l}{\cellcolor[HTML]{EFEFEF}\textit{Audio Flamingo Next}} \\
LLM-eval & 68.0 & 50.0 & 52.0 & 56.7 \\
Human-eval & 68.0 & 52.0 & 52.0 & 57.3 \\
Agreement & 100.0 & 98.0 & 100.0 & 99.3 \\
\bottomrule
\end{tabular}
\end{table}

The LLM judge agrees with human judgments on 297 of the 300 sampled outputs, corresponding to an overall agreement of 99.0\%. All Beyond-Semantic labels agree for both models. The three discrepancies are isolated: two false negatives on Personal Monologues, one for each model, and one false positive on Ultra Multi-Turn for Qwen3-Omni. We observe no systematic subset- or model-specific disagreement in the audited sample.

\subsection{Long-Context Spoken Dialogue Cases in Vox-Infinity}
\begin{figure}[t]
    \vspace{-0.25cm}
    \centering
    \begin{subfigure}[b]{0.32\linewidth}
        \centering
        \includegraphics[width=\linewidth]{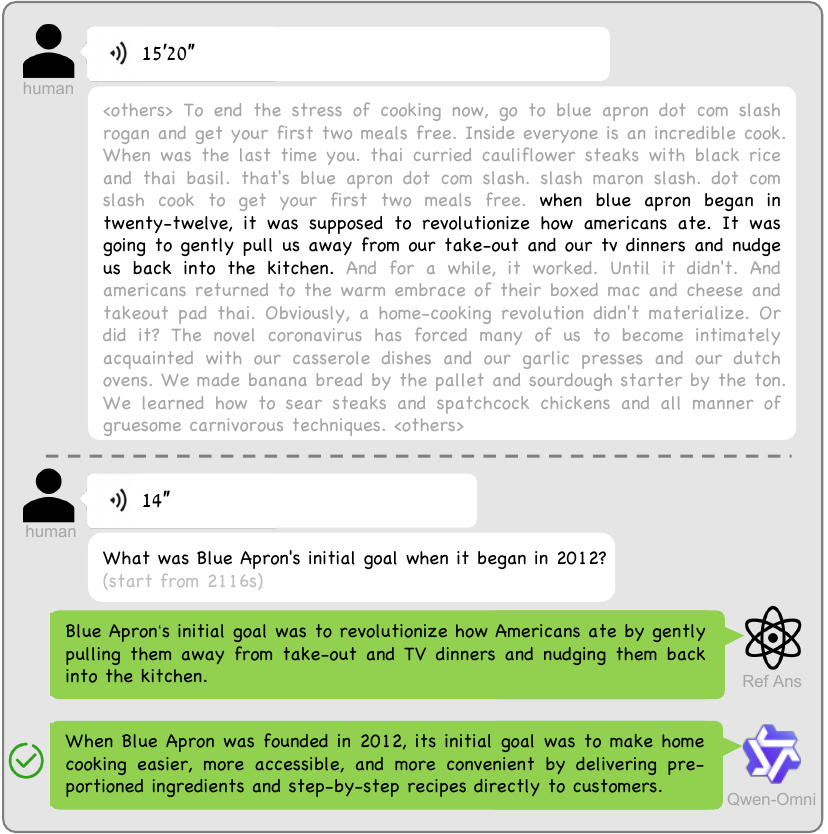}
        \caption{Ultra Multi-Turn Dialogues.}
        \label{fig:a}
    \end{subfigure}
    \hfill
    \begin{subfigure}[b]{0.32\linewidth}
        \centering
        \includegraphics[width=\linewidth]{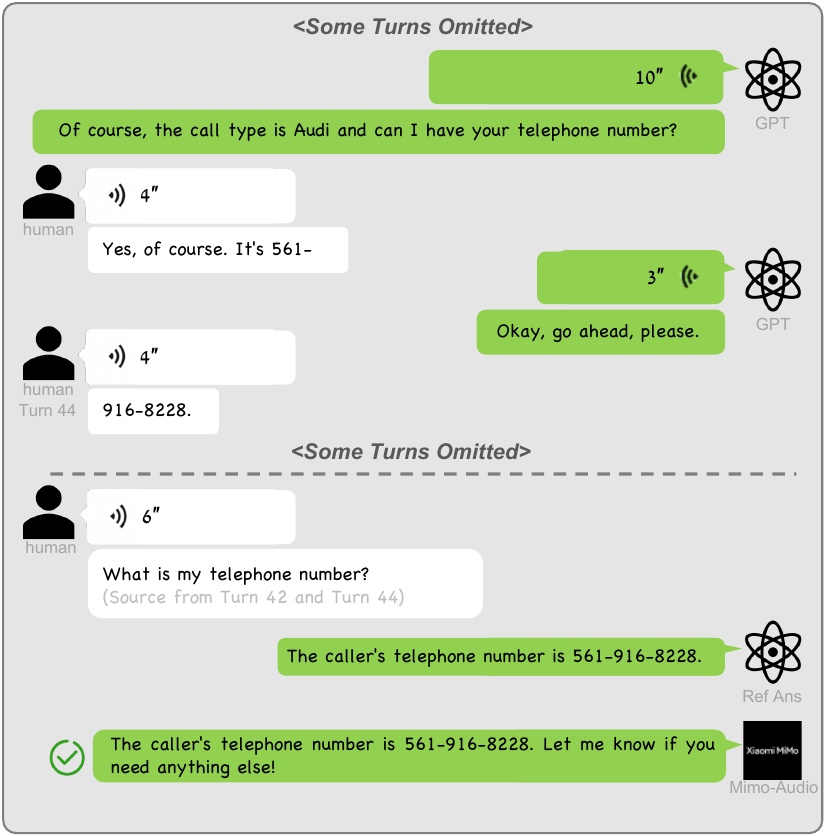}
        \caption{Personal Monologues.}
        \label{fig:b}
    \end{subfigure}
    \hfill
    \begin{subfigure}[b]{0.32\linewidth}
        \centering
        \includegraphics[width=\linewidth]{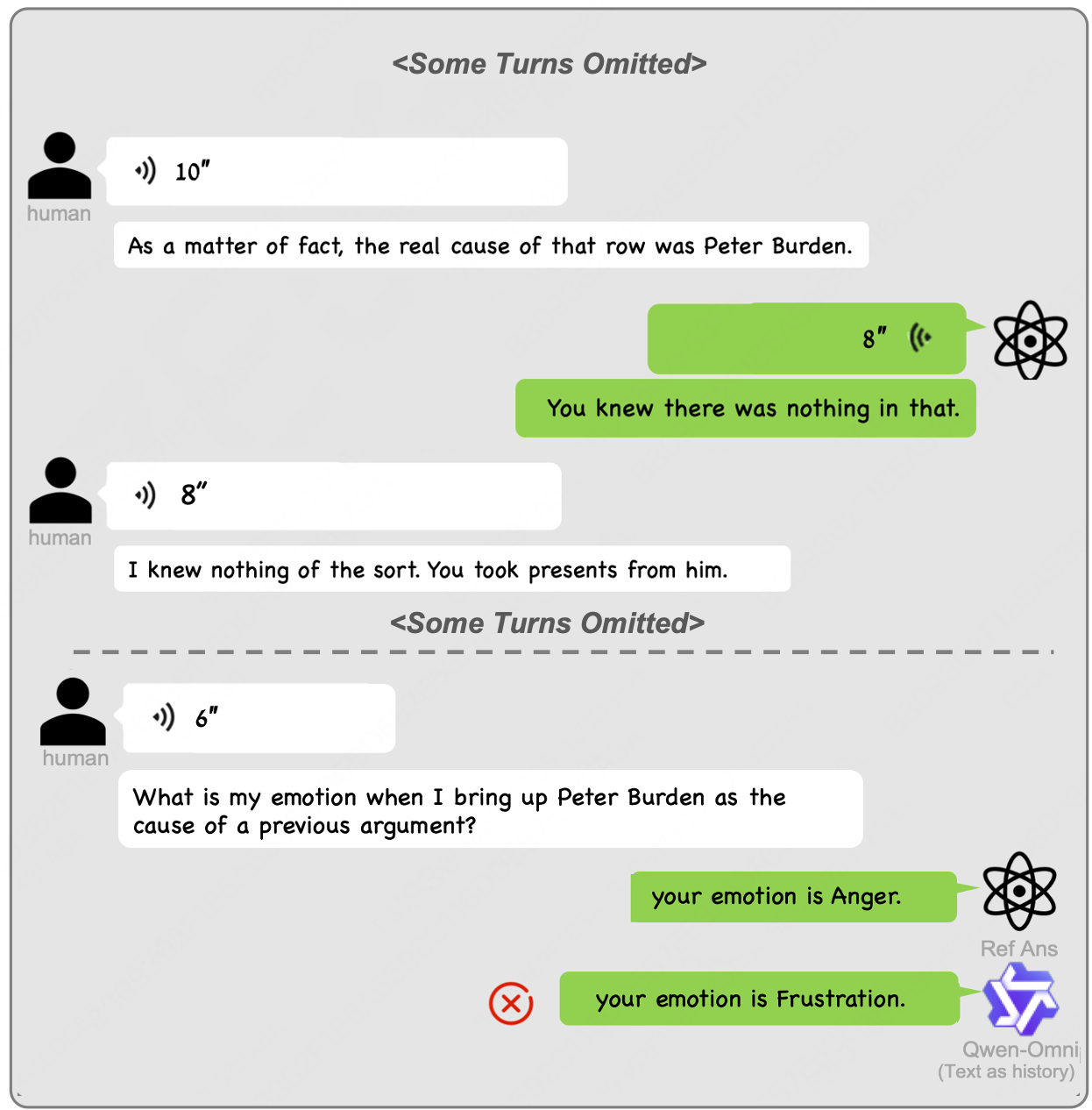}
        \caption{Beyond-Semantic Dialogues.}
        \label{fig:c}
    \end{subfigure}
    \caption{Representative examples drawn from the three subsets of \textsc{Vox-Infinity}.}
    \label{fig:cases}
    \vspace{-0.30cm}
\end{figure}

To illustrate Vox-Infinity, Figure~\ref{fig:cases} presents one example from each subset. In each case, the light-gray region within the question indicates the answer's location, thereby specifying the amount of historical context required. The examples illustrate the distinct evidence patterns targeted by the three subsets. In particular, the Beyond-Semantic example in Figure~\ref{fig:c} shows why text-only histories cannot preserve paralinguistic cues. Additional examples are available on our demo page.

\section{Conclusion}

In this work, we present Vox-Infinity, a benchmark for evaluating long-context understanding in spoken language models. Vox-Infinity extends spoken histories along turn count and turn duration and provides answer-provenance annotations for evidence-distance-aware evaluation across fine-grained retrieval, multi-hop reasoning, and paralinguistic memory. Evaluations of seven representative models reveal an overall recency effect, with model performance generally declining as answer-supporting evidence appears farther from the query. The results further show that fine-grained semantic evidence is difficult to retrieve from long audio histories, while text histories preserve lexical information more effectively but lose speech-specific paralinguistic cues. These findings establish Vox-Infinity as a diagnostic testbed for studying how spoken language models retain and use semantic and acoustic information over extended interactions.

\section*{Ethical Considerations}
Vox-Infinity was developed with careful attention to ethical and responsible research practices. All speech data used in this benchmark are publicly available, and no personally identifiable information is included. To further protect speaker privacy, the data are anonymized and used exclusively for research purposes.

We recognize that spoken dialogue corpora may reflect social, cultural, or gender biases, which could be inherited by models trained or evaluated on them. While Vox-Infinity is intended to advance research on long-context understanding, it should not be regarded as free of bias. We encourage future researchers to critically examine and mitigate potential harms that may arise from evaluation on this benchmark.

Finally, as large language and speech models become increasingly powerful, they also present risks of misuse, such as generating misleading content or enabling surveillance. Vox-Infinity is therefore released solely for academic and responsible industrial research, with the goal of fostering dialogue systems that are beneficial, transparent, and aligned with human values.

\vspace{-0.15cm}
\section*{Reproducibility Statement}

We are committed to ensuring the reproducibility of our work. A portion of the dataset is already publicly available on our demo page, and the full release will be made available after paper acceptance. We provide detailed descriptions of the dataset construction process, evaluation protocols, and experimental settings in the main text and Appendix. In addition, the evaluation tools required for benchmarking will also be released upon acceptance to facilitate replication and future research.

\appendix

\section{Use of LLM}
In this work, we employed large language models (LLMs) both for generating QA data and for conducting model evaluation. Moreover, as Vox-Infinity is proposed as a benchmark, we also report the performance of several LLMs on our dataset to provide reference baselines.

\section{Prompt Template}
\subsection{Prompt Template for QA generation.}
\label{App:QA}
\begin{figure}[H]
    \centering
    \includegraphics[width=0.78\linewidth]{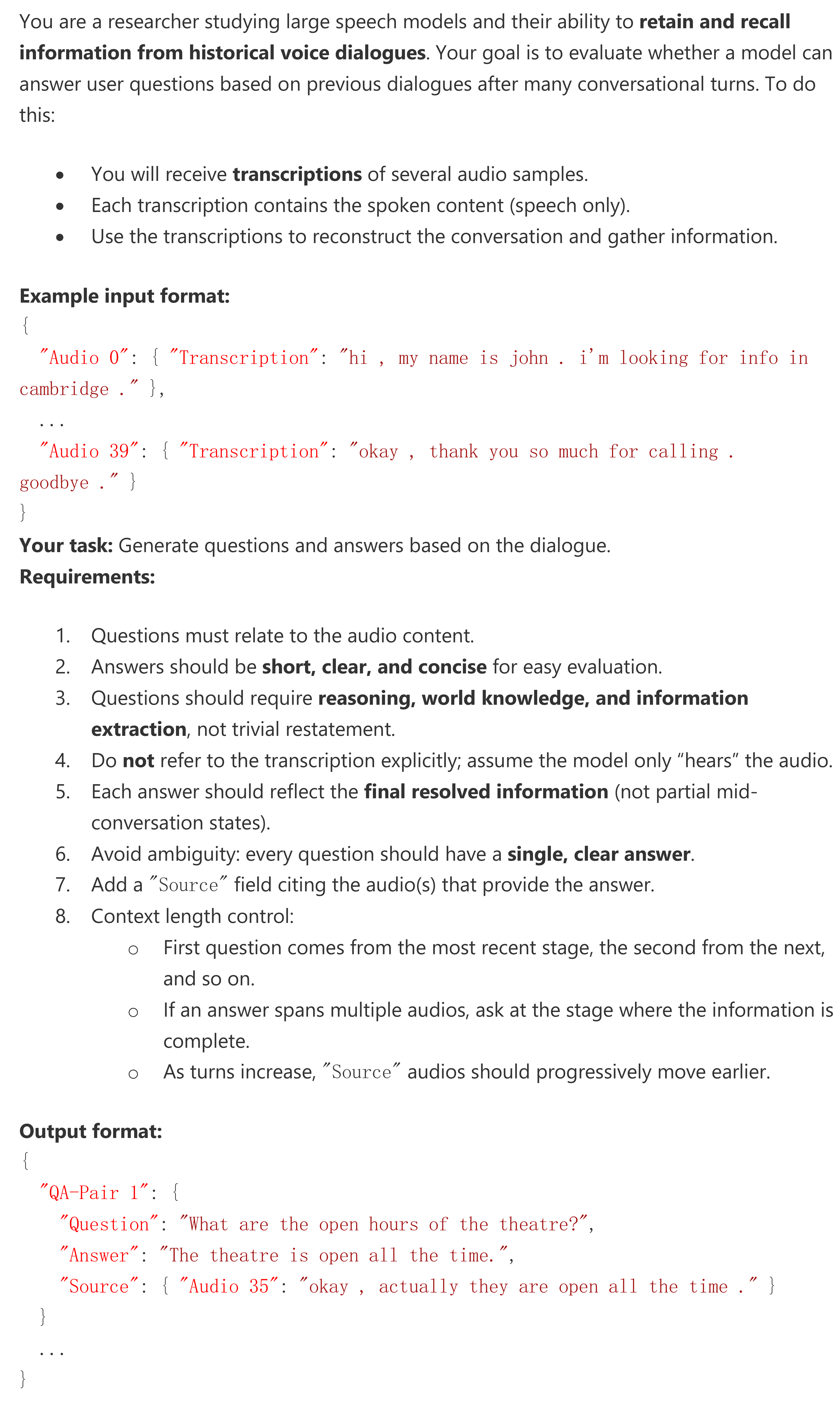}
    \caption{Prompt template for QA generation.}
    \label{fig:QA-prompt}
\end{figure}

We show the prompt template for QA generation in Figure~\ref{fig:QA-prompt}. In addition to generating question–answer pairs, the process also records the source locations of the information needed for answering, ensuring traceability.

\subsection{Evaluation Prompt Template}
\label{app:prompt_template}
Figure~\ref{fig:prompt_template} illustrates the GPT prompt template employed for model evaluation.


\begin{figure}[H]
\centering
\begin{tcolorbox}[colback=white, colframe=black!75,
                  fonttitle=\bfseries, title=Prompt Template,
                  width=0.95\linewidth, boxrule=0.8pt, arc=2pt]
You are an AI assistant tasked with judging the correctness 
of a model's response to a multiple-choice question. 
You are given the question, its answer options, the correct answer, 
and the model's output. Your evaluation must follow these rules:
\\[1em]
- If the model's output matches the correct answer, output ``Correct''. \\ 
- If the model's output matches one of the options but not 
  the correct answer, output ``Incorrect''.  \\
- If the model's output does not match any of the options, 
  output ``Unknown''.  
\\[1em]
Your output must be a single word chosen strictly from:  
\{Correct, Incorrect, Unknown\}.  
\\[1em]
\textbf{Question:} \{question\}  \\
\textbf{Options:} \{options\}  \\
\textbf{Correct Answer:} \{correct\_answer\}  \\
\textbf{Model Output:} \{model\_output\}  \\
\end{tcolorbox}
\caption{Standardized prompt template used for QA evaluation.}
\label{fig:prompt_template}
\end{figure}

\bibliography{references}
\bibliographystyle{iclr2027_conference}

\end{document}